# A Robotic Framework to Facilitate Sensory Experiences for Children with Autism Spectrum Disorder: A Preliminary Study

HIFZA JAVED, George Washington University
RACHAEL BURNS, George Washington University
MYOUNGHOON JEON, Michigan Technological University
AYANNA M. HOWARD, Georgia Institute of Technology
CHUNG HYUK PARK, George Washington University

The diagnosis of Autism Spectrum Disorder (ASD) in children is commonly accompanied by a diagnosis of sensory processing disorders as well. Abnormalities are usually reported in multiple sensory processing domains, showing a higher prevalence of unusual responses, particularly to tactile, auditory and visual stimuli. This paper discusses a novel robot-based framework designed to target sensory difficulties faced by children with ASD in a controlled setting. The setup consists of a number of sensory stations, together with robotic agents that navigate the stations and interact with the stimuli as they are presented. These stimuli are designed to resemble real world scenarios that form a common part of one's everyday experiences. Given the strong interest of children with ASD in technology in general and robots in particular, we attempt to utilize our robotic platform to demonstrate socially acceptable responses to the stimuli in an interactive, pedagogical setting that encourages the child's social, motor and vocal skills, while providing a diverse sensory experience. A preliminary user study was conducted to evaluate the efficacy of the proposed framework, with a total of 18 participants (5 with ASD and 13 typically developing) between the ages of 4 and 12 years. We describe our methods of data collection, coding of video data and the analysis of the results obtained from the study. We also discuss the limitations of the current work and detail our plans for the future work to improve the validity of the obtained results.

CCS Concepts: • Computer systems organization → Robotics;

KEYWORDS

Human-robot interaction, sensory processing disorder, autism spectrum disorder, socially assistive robots

## 1. INTRODUCTION

Sensory abnormalities are reported to be central to the autistic experience. Anecdotal accounts [1-2] and clinical research [3-6] both provide sufficient evidence to support this notion. One study found that, in a sample size of 200, over 90% of children with Autism Spectrum Disorder (ASD) had sensory abnormalities and showed symptoms in multiple sensory processing domains [4]. The symptoms include hyposensitivity, hypersensitivity, multichannel receptivity, processing difficulties and sensory overload. A higher prevalence of unusual responses, particularly to tactile, auditory and visual stimuli, is seen in children with ASD, when compared to their typically developing (TD) and developmentally delayed counterparts. The distress caused by some sensory stimuli can cause self-injurious and aggressive behaviors in children who may be unable to communicate their anguish [7]. Families also report that difficulties with sensory processing and integration can restrict participation in everyday activities, resulting in social isolation for them and their child [8] and impact social engagement [9-11].

Recent research efforts of our team have focused on developing robot-based tools to provide socio-emotional engagement to children with ASD. In our previous work, we developed an interactive robotic framework that included emotion-based robotic gestures and facial expressions to encourage music-based socio-emotional engagement [12-13]. The current work is an extension of these efforts where we use this framework in a pedagogical setting in which the robots model appropriate social responses to salient sensory stimuli that are a part of the everyday experience. This is done in a manner that is interactive and inclusive of the child, such that the robot and the child engage in a shared sensory experience. These stimuli are designed to resemble real world scenarios that form a typical part of one's everyday experiences, such as uncontrolled sounds and light in a public space (e.g. a mall or a park), or tactile contact with clothing made of fabrics with different textures. These are only some of the everyday sensory stimuli that are used as inspiration to design the sensory stations in our setup for this study. Given the strong interest of children with ASD in technology in general and robots in particular [14], we attempt to utilize these robotic platforms to demonstrate socially acceptable responses to such stimulation in an interactive setting that encourages the child to become more receptive to experiences involving sensory stimulation.

With long-term exposure to the robots in this setting, we hypothesize that children will also be able to learn to communicate their feelings about discomforting sensory stimulation as modeled by the robots, instead of allowing uncomfortable experiences to escalate into extreme negative reactions, such as tantrums or meltdowns. However, since the targeted pedagogical effects can only be evaluated through a long-term study comprising of multiple sessions with the participants over a longer period of time, the current paper focuses only on our preliminary evaluation of this framework as a tool to effectively engage the participants as well as model behaviors in ways that are easily interpreted by the participants. Therefore, the analysis and results presented in this paper relate exclusively to the socioemotional engagement of the participants while interacting with the robots, in order to evaluate the capability of our setup to engage the children. In the future, as we invite the same participants for further sessions, the collected data will enable us to evaluate the pedagogical efficacy of our framework as well.

Section 2 of this paper discusses previous studies that have focused on the alleviation of sensory processing difficulties in children with ASD and Section 3 describes the framework and its components in detail. Section 4 describes the preliminary user study, Section 5 explains the various assessment methods employed for the study and Section 6 discusses the metrics derived from the data analysis. The results are discussed in Section 7, and a discussion on these findings is given in Section 8. Section 9 presents a conclusion for this study.

## 2. BACKGROUND

Interventions to address sensory processing difficulties are among the services most often requested by parents of children with ASD [15-16]. The most popular of these is the combination of occupational therapy and sensory integration (OT/SI) [17-18]. However, until recently, there was little support for evidence-based research to support the practice of sensory integration, with the evidence in its favor being largely anecdotal. One of the first rigorous studies in this domain was conducted only recently in 2013 by Schaaf et. al., which found that children who received OT/SI therapy showed a reduced need for caregiver assistance in self-care and socialization [19].

In addition to conventional therapy methods, technologically enhanced intervention methods have also been widely adopted in ASD therapy due to the inherent interest in technology that is commonly reported in children with ASD [20-25]. The range of solutions is diverse, including video-based instruction, computer-aided instruction, mobile applications, virtual reality, and socially assistive robots [13][26-27]. However, such methods mostly target the core impairments that characterize ASD, with applications that focus on improving educational performances, emotional recognition and expression, social behaviors, and/or language and communication skills [28-30]. Relatively fewer research efforts have targeted improvements in sensory processing capabilities.

The MEDIATE (Multisensory Environment Design for an Interface between Autistic and Typical Expressiveness) project uses a multi-sensory, responsive environment designed to stimulate interaction and expression through visual, aural and vibrotactile means [31-32]. It is an adaptive environment that generates real-time stimuli such that low functioning children with ASD, who have no verbal communication, can get a chance to play, explore, and be creative in a controllable and safe space [33].

Several haptic interfaces have also been designed to facilitate therapy and provide non-invasive treatment alternatives [34]. These use vibrotactile, pneumatic, and heat pump actuation devices, designed to be worn on the arm, wrist, leg and chest areas [35]. These devices simulate touch by compressing to provide distributed pressure and emphasize the importance of tactile interactions for mental and emotional health. A vibrotactile gamepad has also been designed that allowed users to receive vibration patterns on the gamepad as they play video games [36]. This vibrotactile feedback corresponds to the events in the videogame but rather than alleviating sensory difficulties, it is primarily meant to convey emotions through the combination of sounds and vibration patterns in attempt to enhance the emotional competence of children with ASD.

In addition, a popular humanoid robot, KASPAR, has also been equipped with tactile sensors on its cheeks, torso, both arms, palms and at the back of the hands and feet [37-38] to test whether appropriate physical social interactions can be taught to children with ASD. It was shown that the robot, by providing appropriate feedback to tactile contact from the children, such as exclaiming when pinched or grabbed by the children, was able to train them to modulate the force they used while touching others [39].

Given the lack of robot-based studies targeting sensory processing difficulties and the mounting evidence emphasizing the need for sensory integration, we have directed our efforts towards this domain. As already stated, our team previously developed an interactive robotic framework that includes emotion-based robotic gestures and facial expressions, which are used to generate appropriate emotional and social behaviors for multi-sensory therapy [12][40]. However, the current study is a step forward in evaluating this design through a preliminary user study that assesses the ability of the framework to maintain children's engagement and to model appropriate responses to sensory stimulation that are easily understood by the children. The following sections will explain these in detail.

3. THE FRAMEWORK

3.1 Design Considerations

When designing the interaction between the robots and the children, we took under consideration several factors pertaining to the appeal of the interaction, the purpose of the activity and the limitations imposed by sensitivity of the children toward sensory stimulation and the capabilities of the robots. Table 1 summarizes these design considerations.

Table 1. Design considerations for child-robot interaction design

| Design consideration | Description |
| --- | --- |
| Choice of robots | The robots must not be too large in size in order to prevent children from being intimidated by them. They must also be capable of expressing emotions through different modalities such as facial expressions, gestures and speech. The robots must also be friendly in order to form a rapport with the children. |
| Appeal of the interaction | The activity being conducted must also be able to maintain a child's interest through the entire length of the interaction. This implies that the duration of the activity, as well as the content must be appealing to the children. |
| Choice of scenarios for sensory stimulation | The sensory station scenarios must be designed to be relatable to the children such that they are able to draw the connection between the stimulation presented to the robots and that experienced by them in their everyday lives. |
| Interpretability of robot actions | The robot actions must be simple and easy to understand for children in the target age range. The gestures, speech, facial expressions and body language must be combined to form meaningful and easily interpretable behaviors. |
| Choice of emotions expressed by robots | The emotion library of the robots must be large enough to effectively convey different reactions to the stimulation but also simple enough to be easily understood by the children. |

To address these concerns, we first chose two small robots: a humanoid robot and an iPod-based robot with a custom-designed penguin avatar. The humanoid robot was programmed to express emotions mostly through gestures and speech, while the iPod-based robot mainly used facial expressions, paired with sound effects and movements allowed by the treads on which it was mounted. Details on these two platforms are given in Section 3.2.

In general, the activity was designed to last between 8-10 minutes per robot, unless the children required or specifically asked for certain actions to be re-performed, in which case it could take longer. The interactions were also designed to be interactive, with the robot directly addressing the children and explicitly requesting their participation. In this manner, the activity was inclusive of the children and ensured that they shared this sensory experience with the robots instead of having to merely observe their behaviors.

The station scenarios used in this interaction were inspired by everyday experiences, in order to expose the child to common sensory stimuli, some indirectly (from observing the robots) and some directly (when robots explicitly ask for the child's participation). The stimuli are designed to be appealing and not overwhelming for the children. However, it was ensured that immediate adjustments to modify the stimuli were possible in case a child had a lower threshold for sensory stimulation than estimated. For example, the volume of the music could be lowered, and the direction of the light beam could be changed in real-time as required.

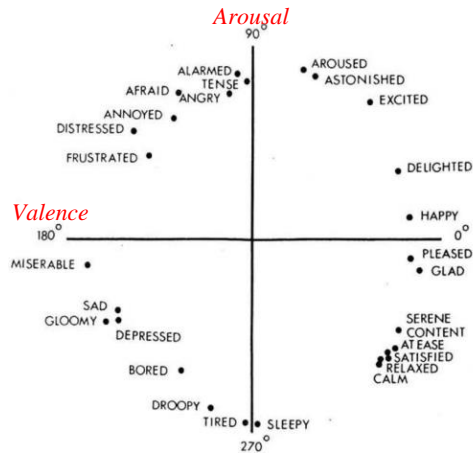

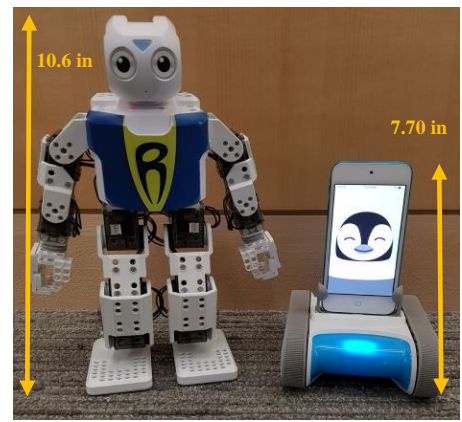

Fig. 1. Russell's Circumplex model of emotion [42]                    Fig. 2. Mini (L) and Romo (R)

The actions of the robots were also designed to depict simple but meaningful behaviors, combining all available modalities of emotional expression, such as movement, speech and facial expressions. The responses of the robots were designed to be expressive, clear and straightforward so as to facilitate interpretation in the context of the scenario being presented at the given sensory station.

The emotion library of the iPod-based robot spanned 20 different emotion states. This was a part of our previous work [41], the goal of which was to develop emotionally expressive robotic agents. These emotions were represented along a valence-arousal scale, as determined by Russell's Circumplex model [42] shown in Fig. 1. These were chosen to incorporate the 6 basic emotions [43], along with other emotion states we deemed most relevant in the context of an interaction involving sensory stimulation. The above applied mainly to the iPod-based robot, which responded to the stimuli with combinations of these emotions through facial expressions and accompanying sound effects. For the humanoid robot, the reactions comprised mostly of gestures and speech, due to which explicit emotional expressions, though programmed into the robot, were not directly utilized. Additional details on the emotional capabilities of each robot are presented in Section 3.2.

3.2 Robotic Platforms

As mentioned, two different robots were employed for this study: the humanoid robot, Robotis Mini (from Robotis) and the iPod-based robot, Romo (from Romotive). The two robots and their heights in inches are shown in Fig. 2. Since Mini communicated through gestures and speech, its responses to the sensory stimulation resembled natural human-human communication more closely than those of Mini, which used relatively primitive means of communication, like facial expressions, sound effects and movements. Therefore, the reactions from Romo comprised of explicit emotional expressions joined one after another to form meaningful responses, whereas Mini could express its responses meaningfully without acting out explicit emotions. In addition to the routines that both the robots were programmed to perform at each of the 5 sensory stations, both robots were capable of expressing individual emotions as well.

The 20 expressions that were programmed into Romo as animations are as follows (shown in Fig. 3): *neutral, unhappy, sniff, sneeze, happy, excited, curious, wanting, celebrating, bored, sleepy, scared, sad, nervous, frustrated, tired, dizzy, disgust, crying and angry*. It must be noted that although *sneeze, wanting* and *sniff* are not emotions, they were chosen for their utility in portraying certain physical states (such as sickness), as well as for aptly depicting responses to certain stimuli. The animation for each emotion was accompanied with a dedicated background color, as well as complementary changes in the tilt angle of the iPod, and circular or back-and-forth movements of the treads upon which it was mounted in order to further facilitate emotional expression [40][44]. It must be pointed out here that the penguin character and the emotion animations were custom-designed and developed by our team as a part of our previous work [41]. Mini was also programmed for the same emotional expressions, some of which are shown in Fig. 4.

Both the robots were controlled individually through their own interfaces. Mini was controlled via a custom-designed user interface from an Android tablet. Romo's actions could be selected in real time and manually communicated to the server [44]. Based on the selected action, a routine that included both the animations and physical movements was then autonomously executed by Romo. The actions of both robots could be manually selected by an instructor based on the children's responses to these actions, in order to adapt to their needs. For example, the instructor could repeat a robot action if the child was distracted or interrupted by a pressing physical need (such as a runny nose).

### 3.3 Sensory Stations and Setup

The setup was composed of five different stations, presenting unique scenarios to stimulate each of the visual, auditory, olfactory, gustatory and tactile senses. The stations were arranged in a fixed sequence: 1) Seeing Station, 2) Hearing Station, 3) Smelling Station, 4) Tasting Station, and 5) Touching Station. These stations were set on a tabletop, 30 inches wide and 60 inches long, placed in the middle of a room. The distances between the stations were as depicted in Fig. 5. The room had glass walls and some workstations set up along its perimeter.

The stimuli and station setup are described as follows.

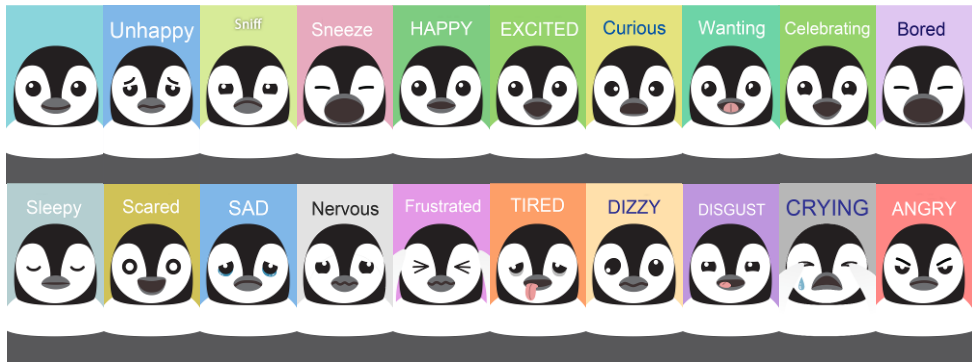

Fig. 3. The 20 emotional expressions of Romo

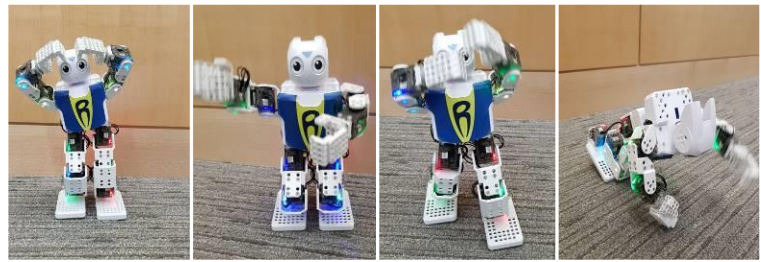

Fig. 4. Emotional expressions of Mini (L-R): dizzy, happy, scared, and frustrated.

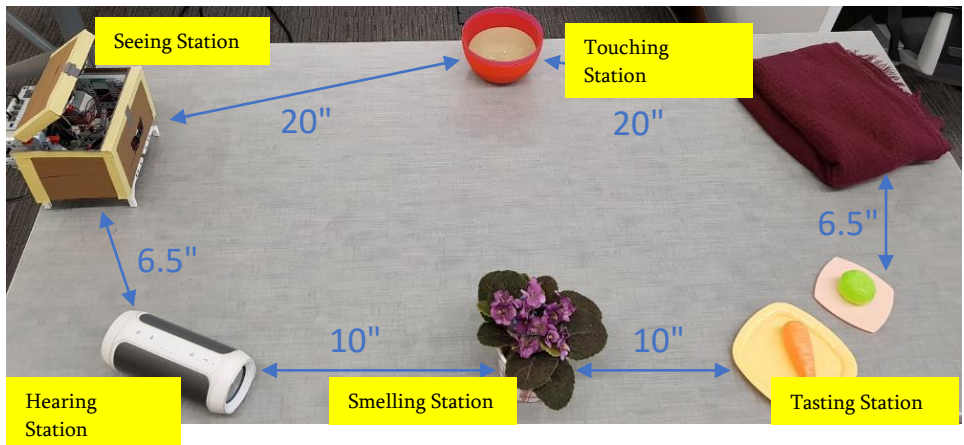

Fig. 5. Setup of the five sensory stations

- Children with ASD often exhibit atypical visual behavior in attempt to either avoid visual input or to seek additional visual stimuli. Therefore, it is common to find them covering their eyes at bright lights or twisting fingers in front of their eyes [7]. At the Seeing Station, we emulated visual stimulus by placing a flashlight inside a lidded box constructed from a LEGO Mindstorm EV3 kit [45]. The front side of the box had an IR sensor that opened the lid when it detected movement in its proximity and the flashlight directed a bright beam of light in the direction of the approaching robot. This formed the visual stimulus at this station.
- Atypical processing of audio stimuli is seen to manifest in the form of unusual behavioral responses commonly observed in children on the autism spectrum [4]. These behaviors include covering of the ears to seemingly benign sounds such as from the vacuum cleaner and the blender. Furthermore, if affected individuals learn to avoid the auditory input that they perceive as unpleasant, it could curtail the learning that comes from listening to the sounds from one's environment [7]. To circumvent this problem, a Hearing Station was designed to provide an auditory stimulus in the form of music played by a Bluetooth speaker. For this study, we played "Llena de Plena" [46], as cheerful dance music to make up the auditory stimulus.
- Though olfactory sensitivity in ASD has not been as widely researched, substantial anecdotal evidence supports heightened olfactory perception in individuals with ASD. An example includes accounts of individuals with ASD refusing to walk on grass because they found the smell overpowering [2]. Experimental reports also suggest heightened olfaction in ASD individuals [47]. In attempt to enable such individuals to become accustomed to smells they are likely to encounter regularly but find overwhelming, a Smelling Station was designed to provide olfactory stimulus in the form of scented artificial flowers inside a flowerpot. The scent could be modified on a case-by-case basis.
- A widely reported attribute of ASD behavior is selective dietary habits [48-49], which can directly impact the nutritional intake in growing children. Factors such as smell, texture, color, and temperature can all contribute to food selectivity. The Tasting Station was designed to encourage diversity in food preferences by presenting a variety of food options. It had two small plastic plates with two different food items that could be modified to match every child's likes and dislikes. These items formed the gustatory stimulus at this station.
- Tactile sensitivity is also commonly reported as a symptom of ASD. Intolerance of certain textures and aversion to certain fabrics is commonly found [50][6], in addition to overreactions to cold, heat, itches or even being touched by other people [51-52]. Therefore, to encourage tolerance of various textures, a soft red blanket and a bowl of sand (with golden stars hidden inside it) were used as the tactile stimuli at the Touching Station.

### 3.4 Five-senses Activity

The sensory scenarios used in this study were designed to closely resemble situations that children would encounter frequently in their everyday lives, hence making them relatable and easy to interpret. Given their strong interest in technology, we attempted to leverage the ability of our robots to elicit a higher level of socio-emotional engagement from these children. The robots are used in this sensory setting to demonstrate socially acceptable responses to stimulation to encourage children to become more receptive to a variety of sensory experiences, as well as to effectively communicate their feelings if the experiences cause them discomfort. As will become clear from the description of the actions of the robots at each sensory station in Table 2, the activity was designed to be interactive, with the robots sharing the sensory experience with the children in order to sustain their engagement. The stations were arranged in the fixed sequence described in Section 3.3 for every session.

The robots show both positive and negative responses at some of the sensory stations with the aim of demonstrating to the children how to communicate their feelings even when experiencing discomforting or unfavorable sensory stimulation, instead of allowing the negative experience to escalate into a tantrum or meltdown. These negative reactions were designed not to be too extreme, so as to focus on the communication of one's feelings rather than encouraging intolerance of the stimulation.

As listed in Table 2, the instructor made sure to pause between each dialog delivered by Mini so as to allow the child to respond before proceeding with the activity. The questions from Mini were meant to promote active participation from the children and keep them interested in the interaction.

It is also important to point out the intended purposes of demonstrated actions of the robots at each station:
- *Seeing Station*: To effectively handle uncomfortable visual stimuli and to communicate discomfort instead of allowing it to manifest as extreme negative reactions (tantrums/meltdowns). This can be especially useful in controlled environments like cinemas and malls where light intensity cannot be fully regulated.

Table 2. Robot behaviors in the five senses game

| **Mini** | **Romo** |
|---|---|
| <td colspan="2" align="center">Greeting/Introduction</td> |  |
| 1. *[waves]* Hello! My name is Mini. What's your name?<br>2. *[with gestures]* It's nice to meet you. It looks like we're in a maze. I wonder what exciting things are going to happen around this maze. Let's start walking and discovering together. Are you ready?<br>3. [starts walking] Let's go! | [No introduction] |
| <td colspan="2" align="center">Seeing Station</td> |  |
| [walks to the lidded box]<br>1. Wow, a treasure chest! I wonder what's inside. What do you think is inside? [walks closer to the box, lid opens up and a flashlight directs a beam of light at Mini]<br>2. Let's find out. Oh, there's a bright light inside the box. But that's okay, we can cover our eyes. *[steps away from the box and shields its face with its hands]* | 1. Approaches the lidded box [lid opens up and a flashlight directs a beam of light at Romo]<br>2. Plays the dizzy/frustrated routine<br>3. Spins away from the light<br>4. Plays the happy animation once it has moved away from the light |
| <td colspan="2" align="center">Hearing Station</td> |  |
| [walks to the speaker]<br>1. It's a radio! I'm feeling shy. I want to dance, but I'm too shy to. *[music plays in background]* Oh, I like this song. I really want to dance to it. *[ties hands back and takes small sideways steps timidly]* I won't be shy if you dance with me. Let's dance! *[music continues playing] [performs a dance routine]*<br>2. I want to see your favorite moves! [music plays again] [waits for child to dance along]<br>3. Wow, that was fun! What is your favorite dance move? *[waits for child to respond]* | 1. Approaches the speaker<br>2. [music plays in background]<br>3. Plays the happy/excited routine with quick to-and-fro movements to simulate dancing |
| <td colspan="2" align="center">Smelling Station</td> |  |
| [walks to the flower pot]<br>1. Oh, flowers! Let's smell the flowers together! *[bends down to smell the flowers]* Ah, ah, ah choo! *[takes a step back]* These flowers made me sneeze. I think they smell good, even though I sneezed. Do they smell good to you? | 1. Slowly approaches the flower pot<br>2. Tilts the display in the direction of the flowers<br>3. Plays the sneeze animation and retreats to convey displeasure |
| <td colspan="2" align="center">Tasting Station</td> |  |
| [walks to the two plates of food]<br>1. I'm hungry for a snack. Let's try some foods. I will eat this first. *[bends down to the food item in the first plate]* Yum, yum! This food tastes delicious. *[rubs its stomach]* I like it! Let's try another one now. *[bends down to the food item in the second plate]* Yuck, yuck! This food tastes bad. *[gestures disgustedly]* I don't like it. Which food do you like the best? *[waits for child to respond]*<br>2. I think I like this food the best. | 1. Approaches the first plate of food<br>2. Plays the nervous/disgust routine, tilts away and retreats from the plate quickly<br>3. Approaches the second plate of food<br>4. Plays the curious/wanting/happy routine with brisk to-and-fro movements to convey pleasure |
| <td colspan="2" align="center">Touching Station</td> |  |
| [walks to the blanket]<br>1. Wow, we've walked over to a blanket. This blanket looks very soft. I think I want to feel it. Can you help feel it with me? *[waits for child to respond]*<br>2. Great! Let's walk over and touch the blanket. *[moves closer] [bends down to touch the blanket and then rubs it]* It feels very soft and fluffy, like a kitten! What soft animal does it remind you of? *[stands back up and waits for child to respond]*<br>[walks over to the bowl of sand]<br>3. *[points to the bowl]* Hey, I think I lost my toys in the sand. Can you help me find them? *[waits for child to retrieve the golden stars]*<br>4. Thank you! | 1. Approaches the blanket<br>2. Moves on top of it<br>3. Pauses and plays the curious/sleepy routine to convey that the blanket is comfortable and that it is taking a nap in it<br>4. [sand bowl not used] |
| <td colspan="2" align="center">Celebration</td> |  |
| 1. Yay! We're all done! Let's celebrate by rolling around! *[performs somersault]* [gets up on its feet, takes a few steps forward and then balances itself on one foot] | 1. Moves to the middle of the table<br>2. Plays the happy/excited/celebration routine, changes tilt angles rapidly and spins around in circles to convey excitement |

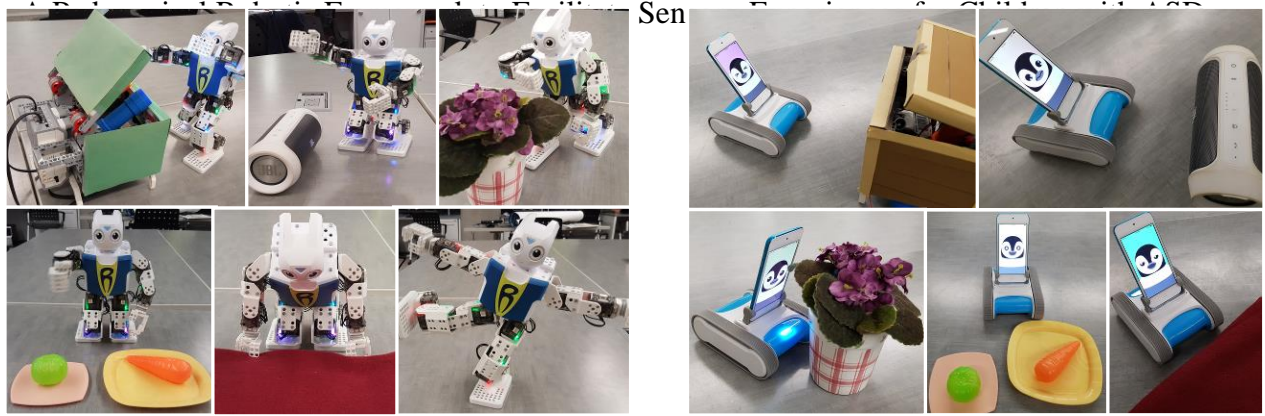

Mini at the five stations.
Top (L-R): Seeing station, Hearing station, Smelling station
Bottom (L-R): Tasting station, Touching station, celebration

Romo at the five stations.
Top (L-R): Seeing station, Hearing station
Bottom (L-R): Smelling station, Tasting station, Touching station

Fig. 6. The robots at sensory stations

- *Hearing Station*: To improve tolerance for sounds louder than those to which one is accustomed, to learn to not be overwhelmed by music, and to promote gross motor movements by encouraging dancing along to it. This can be especially useful in uncontrolled environments like cinemas and malls where sounds cannot be fully regulated.
- *Smelling Station*: To not react with extreme aversion to odors that may be disliked and to communicate the dislike instead. This can be useful for parents of children with ASD who are very particular about the smell of their food, clothes, and/or environments etc.
- *Tasting Station*: To diversify food preferences instead of adhering strictly to the same ones. It must be pointed out that this goal is a long-term target that can only be achieved after the children have already interacted with the robots over several sessions. We anticipate that once the robots have formed a rapport with the child by liking and disliking the same foods as the child, it could start to deviate from those responses, hopefully encouraging the children to be more receptive to the foods their robot "friends" prefer. We plan to change to different food items in the future sessions to achieve this goal. While this remains our eventual target, this claim can only be validated through a long-term study; its mention here is meant only to provide the readers with an insight into the purpose of the station design.
- *Touching Station*: To acclimate oneself to different textures by engaging in tactile interactions with different materials. This is especially useful for those children with ASD who may be sensitive to the texture of their clothing fabrics [50][6] and/or those who experience significant discomfort with wearables like hats and wrist watches etc.

The celebration at the end is meant to convey a sense of shared achievement, while also encouraging the children to practice their motor and vestibular skills by imitating the celebration routines of the robots. Both robots are shown at the sensory stations in Fig. 6.

4. USER STUDY

4.1 Participants

For this study, participants between the ages of 4 and 12 years were recruited. We invited a total of 13 TD and 9 ASD participants. However, we unable to collect data from 2 of the ASD participants due to the severity of their conditions, and data collected from another 2 could not be included since they fell outside our target age range. The TD group consisted of 13 participants, 7 males and 6 females, with a mean age of 7.08 years (SD = 2.565). The ASD group consisted of 5 participants, all males, with a mean age of 8.2 years (SD = 1.095). The demographic details of the participants are listed in Table 3.

4.2 Procedure

We recruited the two groups of participants for multiple sessions of child-robot interactions over a period of a few months, where each child would be allowed to visit maximum of 8 times in total. However, even though we have conducted multiple sessions with some of the participants, this paper presents only the single-session data collected from all participants.

Table 3. Demographic details of the participants

| ID | Age | Gender | Race | Group |
|---|---|---|---|---|
| 1 | 10 | M | Asian | NT |
| 2 | 4 | F | Asian | NT |
| 3 | 5 | F | Asian | NT |
| 4 | 11 | F | White | NT |
| 5 | 9 | M | White | NT |
| 6 | 10 | F | White | NT |
| 7 | 9 | M | White | NT |
| 8 | 5 | M | White | NT |
| 9 | 5 | M | White | NT |
| 10 | 5 | F | White | NT |
| 11 | 5 | M | White | NT |
| 12 | 5 | M | White | NT |
| 13 | 9 | M | South East Asian | NT |
| 14 | 7 | M | White | ASD |
| 15 | 8 | M | White | ASD |
| 16 | 10 | M | White | ASD |
| 17 | 8 | M | White | ASD |
| 18 | 8 | M | South East Asian | ASD |

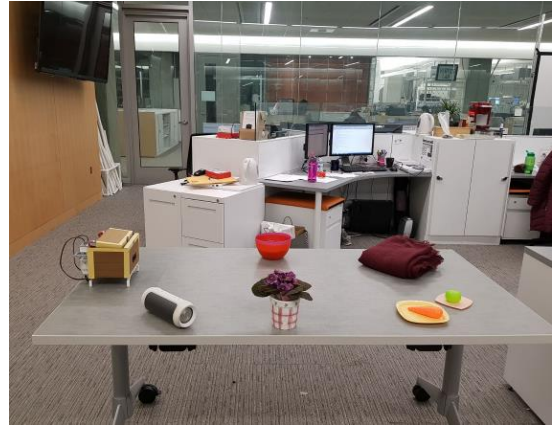

Fig. 7. The setting used to conduct sessions with the participants.

All sessions were conducted individually, not in pairs or groups. The children were accompanied by their parent(s) who remained with them in the room throughout the session. They were given 5-7 minutes to explore the new environment and settle down. The parents were asked to sign the required consent forms and fill out the questionnaires (discussed in Section 5). The sessions were not strictly structured, in that the participants were only briefly instructed about play time with the robots. They were given no explicit directions about how to interact with the robots. They could sit in a chair or remain standing during the session, as they pleased. This method of providing minimal instructions was intentional since it facilitated a naturalistic interaction between the children and the robots. The size of the table and its placement in the room allowed the children to move around freely in order to ensure that they could view the robot from the front at every station (Fig. 7).

In every session, the activity was first conducted with Romo, followed by Mini, and the stations were placed in the same fixed order as described in the previous section. The control mechanisms of the robots and the stimuli design allowed for flexibility, such that an action could be repeated if the interaction was interrupted (for example, if a child had a runny nose or needed to use the restroom) or a stimulus could be adjusted if necessary (for example, the volume of the music could be lowered or increased at the Hearing Station).

A camcorder, mounted on a tripod, was placed at the end of the table, at an angle that ensured that the child and the robots always remained in its view. The footage was used for post-experimental analysis.

## 5. ASSESSMENT METHODS

### 5.1 Video Coding

A behavioral coding software, Behavioral Observation Research Interactive Software (BORIS)[53] the behaviors of interest, as described in Section 5.2. The software was also used to generate behavioral metrics as percentages of total session times for which the respective behaviors were observed.

### 5.2 A Measure of Engagement

In order to derive a meaningful quantitative measure of engagement, we utilized several key behavioral traits of social interactions. These behaviors were selected because they have proven to be useful measures of social attention and social responsiveness from previous studies [54-65]. By coding the video data for these target behaviors through the coding software, we were able to derive an engagement index as the indicator of every child's varying social engagement throughout the interaction with the robots.

These metrics were measured as percentages of the total session time for which the respective behaviors were observed. The engagement index was computed as a sum of these factors, each with the same weight. The engagement index was then normalized to a maximum value of 1. These are described in Table 4.

### 5.3 Questionnaires

As a tool to monitor long-term behavioral changes in the children participating in multiple sessions though the course of the study, we also designed 4 different questionnaires to gather parental reports of their progress. These mostly consisted of close-ended questions with response options on a seven-point Likert scale. The questions were related to the child's general sensory processing abilities, in addition to the parent's report on their interactions with the robots at specific sensory stations. Excerpts from these questionnaires are shown in Table 5.

A baseline assessment questionnaire evaluated the child's sensory perception, as well as language, emotional, social and play skills prior to their participation in this study. This was completed by the parent only once before the participation began. The pre-session questionnaire had more focused questions about the same skills as in the baseline questionnaire and was filled out prior to every session that the child participated in through the course of this study. A post-session questionnaire was filled out upon completion of every individual session. This questionnaire contained questions specifically intended to elicit the parents' feedback on the child's interaction with the robots in that particular session. A post-study questionnaire contained questions related to the child's overall experience through the course of the study. This was required to be filled out only once after all the child's sessions had been completed.

While the pre- and post-session questionnaires monitored the child's interaction with the robots in every individual session, the baseline and post-study questionnaires extracted a more comprehensive picture of the child's overall progress with long-term interactions with our robots. Since we have only conducted single sessions with the participants so far, the analysis presented in this study does not include the data collected from these questionnaires.

## 6. DATA ANALYSIS

### 6.1 Engagement Index and Effectiveness of Robot Actions

The video coding results generated by BORIS contained the timing and behavioral change information for every participant throughout the duration of their respective sessions. These results were run through a MATLAB script to compute the changing engagement index value as the normalized sum of target behaviors observed at regular time intervals throughout the interaction. The staircase plot shown in Fig. 8 depicts the engagement index value for one of the participants. The red plot marks the general engagement trend for this participant. Such plots were generated for all the participants in the study.

The same script also extracted information pertaining to the timing of the robot actions in response to which these engagement index values were generated. The yellow areas in Figure 8 are the periods marking the sensory station actions of the robots. This allowed us to compare the participant's engagement resulting from the actions performed by the robots at the sensory stations versus the engagement resulting from actions when the robot was not at any station. This was used as a method to assess the effectiveness of our design (including sensory stimulation and robot actions at the stations) in engaging children.

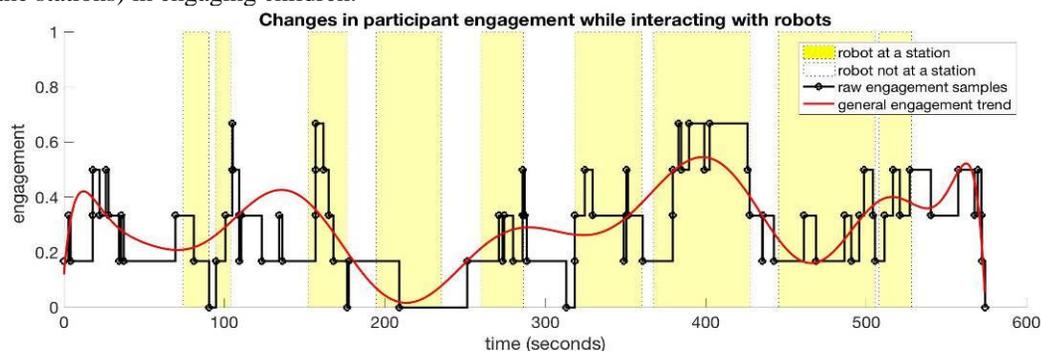

Fig. 8. A plot showing the changes in a participant's engagement during the interaction with robots

Table 4. Behaviors comprising the engagement index

| Behavior | Description |
|---|---|
| Eye gaze focus | Deficits in social attention and establishing eye contact are two of the most commonly reported deficits in children with ASD. We therefore used the children's gaze focus on the robots and/or the setup to mark the presence of this behavior. |
| Imitation | Infants have been found to produce and recognize imitation from the early stages of development, and both these skills have been linked to the development of socio-communicative abilities. In this study, we monitored a child's unprompted imitation of the robot behaviors as a measure of their engagement in the interaction. |
| Vocalizations/ verbalizations | The volubility of utterances produced by children with ASD is low compared to their TD counterparts. Since communication is a core aspect of social responsiveness, the frequency and duration of the vocalizations and verbalizations produced by the children during the interaction is also important in computing the engagement index. |
| Self-initiated interactions | Children with ASD prefer to play alone and make fewer social initiations compared to their peers. Therefore, we recorded the frequency and duration of the interactions with the robot initiated by the children as factors contributing to the engagement index. Examples of self-initiated interactions can include talking to the robots, attempting to feed the robots, guiding the robots to the next station etc. without any prompts from the instructors. |
| Triadic interactions | A triadic relationship involves three agents, including the child, the robot and a third person that may be the parent or the instructor. In this study, the robot acts as tool to elicit interactions between the child and other humans. An example of such interactions is the child sharing her excitement about the dancing robot by directing the parent's attention to it. |
| Smile | Smiling has also been established as an aspect of social responsiveness. We recorded the frequency and duration of smiles displayed by the children while interacting with the robots, as a contributing factor to the engagement index. |

Table 5. Excerpts from questionnaires used to extract parental feedback

| Questionnaire | Sample questions |
|---|---|
| Baseline | 1. Does the child avoid messy play i.e. sand, mud, water, glue, slime, playdoh etc.?<br>2. Is the child frequently distracted by sounds not normally noticed by others (like humming of lights or refrigerators, fans, heaters or clocks ticking)?<br>3. Is the child a picky eater or has extreme food preferences like limited repertoire of foods, picky about brands, resistive to trying new food, or not eating at other people's houses?<br>4. Does the child react negatively to or dislikes smells that do not usually get noticed by or bother other people?<br>5. Is the child easily distracted by other visual stimuli in the room like movements, decorations, windows, doorways etc.? |
| Pre-session | 1. Does the child avoid touching certain textures of material (blankets, rugs, stuffed animals)?<br>2. Is the child fearful of the sound of a flushing toilet, vacuum, hairdryer, squeaky shoes etc.?<br>3. Does the child prefer to eat foods with only certain textures?<br>4. Does the child have difficulty with imitative play? |
| Post-session | 1. Was the child able to understand the robot's dialogs?<br>2. Did the child attempt to share his/her excitement about the robot with you?<br>3. Did the child attempt to do anything during the activity that he/she otherwise finds difficult to do?<br>4. Was the child startled by the robot's movements? |
| Post-study | 1. Did the interaction with robots help the child learn any new skills that may reduce his/her sensory difficulties?<br>2. Has the child attempted to re-create any part of this activity at home or when playing with his/her friends?<br>3. Has the child shown any improvements in vocalization skills since participating in this activity? |

To determine the statistical significance of the differences in the engagement values for the conditions described above, we then conducted a T-test ($\alpha = 0.05$) assuming equal means but unknown variances. This T-test was performed for every participant in both groups for individual analysis of their interaction with the robots.

### 6.2 Inter-group Analysis

To compare the effectiveness of our framework for participants from both groups, we obtained a behavior-wise breakdown of the engagement from the video coding data. This enabled us to compare the performances of both groups with respect to each target behavior. In addition, in order to examine the differences in the engagement of the two groups under the two conditions (i.e. when the robot is at a sensory station versus when the robot is not at any station), the same T-test ($\alpha = 0.05$) was also performed for the grouped data to determine the statistical significance of the differences in the group performances.

### 6.3 Intercoder Reliability Assessment

To improve the reliability of the obtained results, the video coding was completed by three different coders, with the same analysis methods performed on the three sets of generated data. This included both individual and group analysis. The quality of the coding was assessed using intraclass correlation (ICC) of type (3,1), which is a popular method of determining coder's agreement in behavioral sciences [66]. This value ranges between 0 and 1, and according to a guideline provided by Koo and Li [67], values between 0.7-0.9 are considered good, and those above 0.9 are considered excellent. The agreement score and the standard deviation of the engagement data provided by the three coders in this study was $0.8342 \pm 0.159$. ICC values obtained for each participant are obtained from the 3 coders are shown in Fig. 9. All the scores except one (participant 15) achieved a high agreement value of above 0.7.

## 7. RESULTS

### 7.1 Individual Analysis

An engagement index value was computed for every participant's interaction session with the robots, as described in Section 5.1. Using the behavioral metrics obtained from the coding, this engagement index value was computed over the duration of every participant's complete session, as shown in Fig. 10.

To evaluate the efficacy of the designed robot behaviors and sensory scenarios, engagement index values while the robot is at a sensory station were compared with the values obtained when the robot was not at any station. The results of the T-test described in Section 5.1 are given in Table 6. A two-tailed test (tail direction = both) was first performed to establish whether a difference in engagement existed, and one-tailed tests were then performed to determine the direction of the difference. *h-values* of 1 for the left-tailed test show lower engagement for the station actions of the robots, while *h-values* of 1 for the right-tailed test show higher engagement for station actions. As listed in Table 6, the results from the T-test, 5 out of 13 participants in the TD group showed higher engagement while the robot was at a sensory station (ID = 7, 9, 10, 11, 12), 1 participant showed lower engagement (ID = 4), while the remaining 7 showed no difference in their engagement patterns under the two conditions. For the ASD group, 3 out of 5 participants (ID = 14, 16, 17) showed higher engagement while the robot was at a sensory station, and the remaining 2 showed no difference in performances under the two conditions. These results are promising in validating the effectiveness of the robot behaviors designed for the sensory stations.

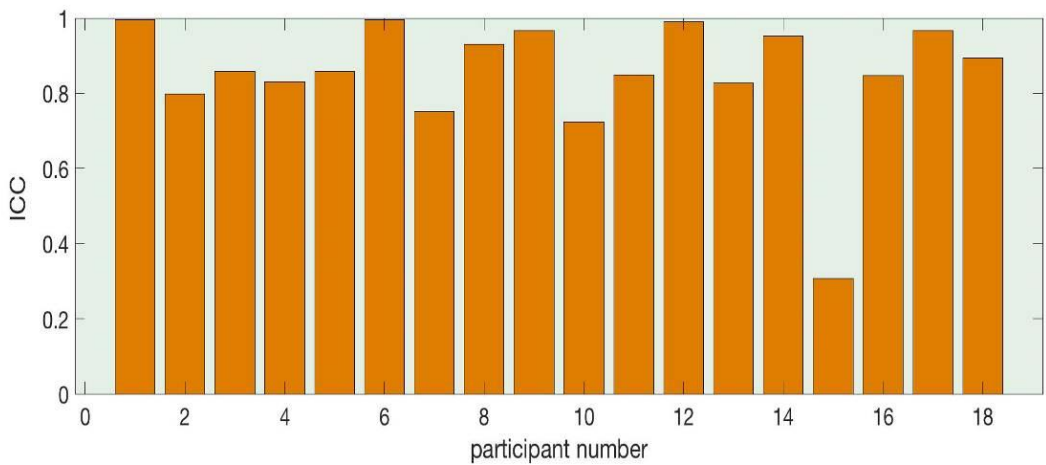

Fig. 9. ICC values obtained per participant from the 3 coders

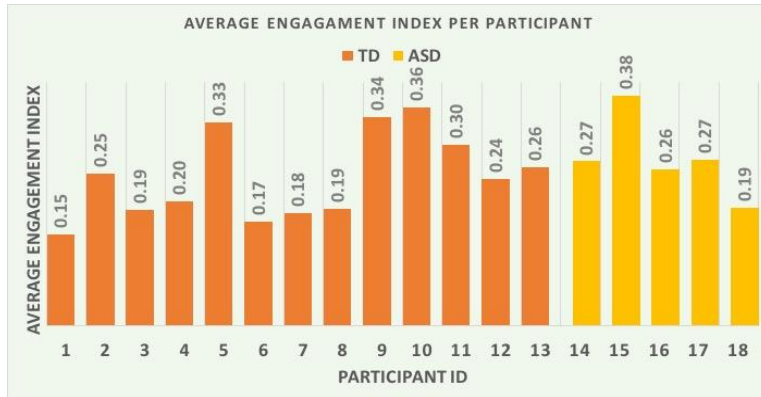

Fig. 10. Engagement indexes for all participants obtained over the lengths of complete sessions

7.2 Inter-group Analysis

To identify any differences in the engagement patterns of the two groups, we first compared the averages of the percent duration of time spent exhibiting each of the target behaviors over complete sessions for the two groups. Though children with ASD performed better for all target behaviors other than gaze, as shown in Fig. 11, these differences were not found to be statistically significant. However, this implies that the overall performances of the two groups with respect to each target behavior were similar, which vouches for the potential of this framework to effectively elicit important target behaviors from children as they interact with our robots such that their performances match those of the TD children.

To compare the overall engagement of the two groups, we used the engagement index values from the time series data of the participants' sessions. A T-test was then performed to identify any differences in the engagement of the two groups for the station and non-station robot actions. As before, a two-tailed test was performed to determine if a difference in the engagements of the two groups existed for each case, and the one-tailed tests were performed to determine the direction of this difference. These results are shown in Table 7. A statistically significant difference in the engagements of the two groups exists for the case when the robot is at a sensory station. An *h-value* of 1 is obtained for the right-tailed test, showing that the ASD group showed a statistically significant higher engagement in the interaction while the robot performed sensory station actions as compared with the TD group. No statistically significant difference in engagement exists for non-station robot behaviors. These results confirm that the designed robot behaviors were effective in engaging the children with ASD, whose performance exceeded that of their TD counterparts.

8. DISCUSSION AND FUTURE WORK

In this work, we designed and implemented a framework for children with comorbid ASD and sensory processing disorders. Though the aim of the overall study is to use this as a pedagogical framework to model and teach socially acceptable responses to a variety of sensory stimuli, the current study focuses only on the evaluation of this framework as a tool for social engagement for children with ASD. This is an essential step toward the evaluation of any robot-based behavioral intervention tool for ASD, since it must be able to sustain the interest of a child and maintain his/her engagement in the interaction for the intervention to have success. To this end, we conducted several tests to analyze both the individual performances, as well as the overall performances of the ASD and TD groups.

The average engagement indices per participant (Fig. 10) show comparable engagement index values for the two groups, with values ranging from 0.15 to 0.36 for the TD group, and 0.19 to 0.38 for the ASD group. To validate the efficacy of the robot behaviors at the sensory stations, we first evaluated the differences in engagement of individual participants while the robot was at a sensory station vs. while the robot was not at any station. In the TD group, 38.46% of participants showed higher engagement with the robot while it was at a sensory station, 7.69% showed lower engagement, and 53.85% showed no statistically significant difference in the engagement level under the 2 conditions.

Table 6. Results of the T-test to determine differences in engagement for station vs. non-station robot actions

| ID | Group | Tail direction for T-test | | | | | |
|---|---|---|---|---|---|---|---|
| | | Left | | Right | | Both | |
| | | p-value | h-value | p-value | h-value | p-value | h-value |
| 1 | TD | 0.879166 | 0 | 0.120834 | 0 | 0.2417000 | 0 |
| 2 | TD | 0.247745 | 0 | 0.752255 | 0 | 0.4955000 | 0 |
| 3 | TD | 0.813215 | 0 | 0.186785 | 0 | 0.3736000 | 0 |
| 4 | TD | 0.000029 | 1 | 0.999971 | 0 | 0.0000572 | 1 |
| 5 | TD | 0.834854 | 0 | 0.165146 | 0 | 0.3303000 | 0 |
| 6 | TD | 0.833057 | 0 | 0.166943 | 0 | 0.3339000 | 0 |
| 7 | TD | 0.992587 | 0 | 0.007413 | 1 | 0.0148000 | 1 |
| 8 | TD | 0.730917 | 0 | 0.269083 | 0 | 0.5382000 | 0 |
| 9 | TD | 0.979207 | 0 | 0.020793 | 1 | 0.0416000 | 1 |
| 10 | TD | 0.998845 | 0 | 0.001155 | 1 | 0.0023000 | 1 |
| 11 | TD | 0.999763 | 0 | 0.000237 | 1 | 0.0018000 | 1 |
| 12 | TD | 0.999122 | 1 | 0.000878 | 1 | 0.0004750 | 1 |
| 13 | TD | 0.84482 | 0 | 0.15518 | 0 | 0.3103590 | 0 |
| 14 | ASD | 0.999923 | 0 | 0.000077 | 1 | 0.0001550 | 1 |
| 15 | ASD | 0.692111 | 0 | 0.307889 | 0 | 0.6158000 | 0 |
| 16 | ASD | 0.978843 | 0 | 0.021157 | 1 | 0.0423000 | 1 |
| 17 | ASD | 0.994237 | 0 | 0.005763 | 1 | 0.0115000 | 1 |
| 18 | ASD | 0.939436 | 0 | 0.060564 | 0 | 0.1211280 | 0 |

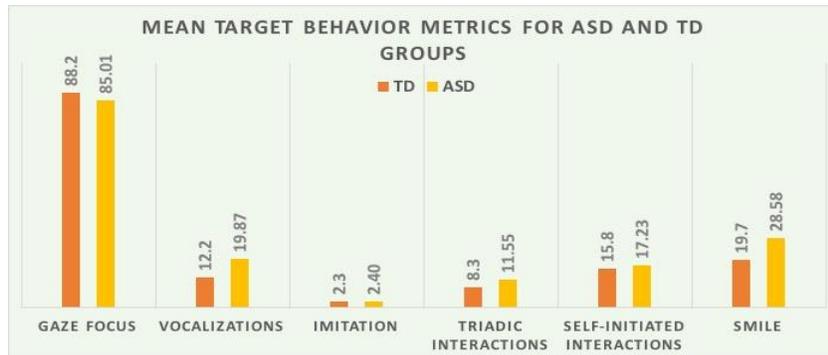

Fig. 11. Average percentage of time each target behavior was exhibited by the two groups

Table 7. Comparing the inter-group engagements for the two different conditions of robot behaviors

| | Tail direction for T-test | | | | | |
|---|---|---|---|---|---|---|
| | Left | | Right | | Both | |
| | p-value | h-value | p-value | h-value | p-value | h-value |
| Robot at a station | 0.9918 | 0 | 0.0082 | 1 | 0.0163 | 1 |
| Robot not at a station | 0.8800 | 0 | 0.1200 | 0 | 0.2400 | 0 |

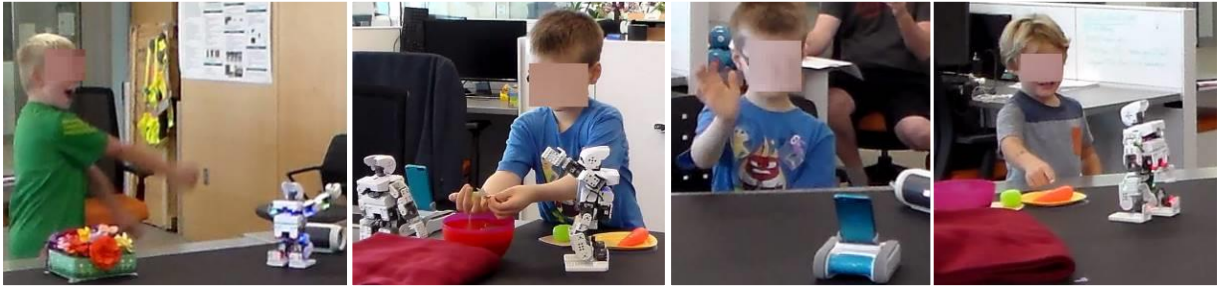

Fig. 12 Participants a) dancing with the robot, b) following instructions from the robot, c) waving at the robot and d) guiding the robot

On the other hand, 60% of participants in the ASD group showed higher engagement while the robot was at a sensory station and the remaining 40% showed no statistically significant difference in engagement under the two conditions. Therefore, in either group, there was only one case of lowered engagement with the sensory station responses of the robots. All other participants showed either an increased engagement or the same level of engagement with the sensory responses of the robots, which is a promising result that confirms the validity of the designed robot behaviors in providing social engagement to the children.

The results of the inter-group analysis also confirm this finding. The effectiveness of the robot behaviors at the sensory stations was tested for the two groups, and it was found that the ASD group showed a higher overall engagement with the robots at the sensory stations as compared with the TD group. For non-station robot actions, both groups had similar performances. In addition, no statistically significant difference was seen in the overall behaviors of the two groups with respect to each of the six target behaviors, which implies that the ASD group showed similar levels of gaze, vocalization, imitation, smiling, and triadic and self-initiated interactions during their interactions with the robots. This is also a promising result that shows that this framework enables children with ASD to match the performance of their TD peers with respect to these key behavioral features of social interactions.

Though children with ASD are typically known to lack social engagement, while interacting with our robots in the current framework, we found them to show either similar or higher social engagement than the TD participants in all respects. We, therefore, conclude that this framework is indeed a suitable tool to provide social engagement to children with ASD and that the robot behaviors at the sensory stations contribute to the achievement of this goal. This also validates that this framework can potentially be useful as an intervention tool for sensory processing disorders. Achieving this, however, is the goal of our next study.

It is important to highlight some interesting observations that were made during the child-robot interactions during the course of this study. Firstly, instances of imitation were mostly observed at the Hearing station (Fig. 12a), which emphasizes the usefulness of incorporating music in child-robot interactions to encourage increased motor and imitative functions. Secondly, the participants were observed to show a good understanding of the actions and intentions of the robots. This was most evident at the Touching station where the participants from both groups followed the request of the robot to find its toys in the sand without waiting for a prompt or permission from the instructor (Fig. 12b). Thirdly, there were several instances where the participants treated the robots as real social entities, as shown in Fig. 12c and 12d where the participants were observed to guide the robot to the next station and wave at it as it turned to face them respectively. These observations point to the potential of socially assistive robots as effective intervention tools for ASD due to their ability to engage children in social interactions.

As a next step, we plan to test the pedagogical capabilities of this framework within the same sensory setup to evaluate its effectiveness as an intervention tool for ASD and sensory processing disorders. This will utilize the data collected through the questionnaires to monitor the participant's long-term progress over the course of the study. In addition, we are also working on the automated assessment of a child's social engagement during an interaction with the robots by using a multi-modality sensing and perception system that utilizes body movement, speech, facial expression, and physiological signal analyses using unobtrusive sensors in order to estimate the internal state of a child. This will allow us to expand the capabilities of the current framework by modulating the robot behaviors in real-time based on the multi-modality behavioral and physiological data collected from the child, and in doing so, take one step closer to the simulation of naturalistic social interactions through robots.

## 9. CONCLUSION

Sensory processing difficulties are commonly experienced by children with ASD and have been found to limit their social experiences, often resulting in social isolation for both the children and their families. While socially assistive robots are being used extensively to target core ASD deficits, very few research efforts have targeted sensory processing disorders. In this study, we designed a novel framework that uses two different robotic platforms to socially engage children with ASD while modeling appropriate responses to common sensory stimuli. With the preliminary user study presented in this paper, we have established the capabilities of this framework as an effective tool for social engagement and hence, a potential tool for behavioral interventions for ASD. As our next step, we will be conducting a large-scale, longitudinal user study to evaluate this framework as an intervention tool for sensory processing difficulties that are comorbid with ASD.

A Pedagogical Robotic Framework to Facilitate Sensory Experiences for Children with ASD   1:18[54] Dubey, Indu, Danielle Ropar, and Antonia F. de C Hamilton. "Measuring the value of social engagement in adults with and without autism." *Molecular autism* 6, no. 1 (2015): 35.
[55] Sanefuji, Wakako, and Hidehiro Ohgami. "Imitative behaviors facilitate communicative gaze in children with autism." *Infant Mental Health Journal* 32, no. 1 (2011): 134-142.
[56] Ingersoll, Brooke. "The social role of imitation in autism: Implications for the treatment of imitation deficits." *Infants & Young Children* 21, no. 2 (2008): 107-119.
[57] Tiegerman, Ellenmorris, and Louis H. Primavera. "Imitating the autistic child: Facilitating communicative gaze behavior." *Journal of autism and developmental disorders* 14, no. 1 (1984): 27-38.
[58] Slaughter, Virginia, and Su Sen Ong. "Social behaviors increase more when children with ASD are imitated by their mother vs. an unfamiliar adult." *Autism Research* 7, no. 5 (2014): 582-589.
[59] Tiegerman, Ellenmorris, and Louis Primavera. "Object manipulation: An interactional strategy with autistic children." *Journal of Autism and Developmental Disorders* 11, no. 4 (1982): 427-438.
[60] Katagiri, Masatoshi, Naoko Inada, and Yoko Kamio. "Mirroring effect in 2-and 3-year-olds with autism spectrum disorder." *Research in Autism Spectrum Disorders* 4, no. 3 (2010): 474-478.
[61] Contaldo, Annarita, Costanza Colombi, Antonio Narzisi, and Filippo Muratori. "The social effect of "being imitated" in children with autism spectrum disorder." *Frontiers in psychology* 7 (2016): 726.
[62] Stanton, Cady M., Peter H. Kahn Jr, Rachel L. Severson, Jolina H. Ruckert, and Brian T. Gill. "Robotic animals might aid in the social development of children with autism." In *Proceedings of the 3rd ACM/IEEE international conference on Human robot interaction*, pp. 271-278. ACM, 2008.
[63] Tapus, Adriana, Andreea Peca, Amir Aly, Cristina Pop, Lavinia Jisa, Sebastian Pintea, Alina S. Rusu, and Daniel O. David. "Children with autism social engagement in interaction with Nao, an imitative robot: A series of single case experiments." *Interaction studies* 13, no. 3 (2012): 315-347.
[64] Wimpory, Dawn C., R. Peter Hobson, J. Mark G. Williams, and Susan Nash. "Are infants with autism socially engaged? A study of recent retrospective parental reports." *Journal of Autism and Developmental Disorders* 30, no. 6 (2000): 525-536.
[65] Nadel, Jacqueline. "Imitation and imitation recognition: Functional use in preverbal infants and nonverbal children with autism." *The imitative mind: Development, evolution, and brain bases* 4262 (2002).
[66] Shrout, Patrick E., and Joseph L. Fleiss. "Intraclass correlations: uses in assessing rater reliability." *Psychological bulletin* 86, no. 2 (1979): 420.
[67] Koo, Terry K., and Mae Y. Li. "A guideline of selecting and reporting intraclass correlation coefficients for reliability research." *Journal of chiropractic medicine* 15, no. 2 (2016): 155-163.